\documentclass[pdflatex,sn-mathphys-num]{sn-jnl}% Math and Physical Sciences Numbered Reference Style 
\usepackage{graphicx}%
\usepackage{multirow}%
\usepackage{amsmath,amssymb,amsfonts}%
\usepackage{amsthm}%
\usepackage{mathrsfs}%
\usepackage[title]{appendix}%
\usepackage{xcolor}%
\usepackage{textcomp}%
\usepackage{manyfoot}%
\usepackage{booktabs}%
\usepackage{algorithm}%
\usepackage{algorithmicx}%
\usepackage{algpseudocode}%
\usepackage{listings}%
\usepackage{amssymb} % For \checkmark
\usepackage{pifont} % For \xmark
\newcommand{\xmark}{\ding{55}} % Define the xmark symbol
\theoremstyle{thmstyleone}%
\theoremstyle{thmstyletwo}%

\theoremstyle{thmstylethree}%

\begin{document}

\title[Article Title]{Multi-Modal Parameter-Efficient Fine-tuning via Graph Neural Network}

%%=============================================================%%
%% GivenName	-> \fnm{Joergen W.}
%% Particle	-> \spfx{van der} -> surname prefix
%% FamilyName	-> \sur{Ploeg}
%% Suffix	-> \sfx{IV}
%% \author*[1,2]{\fnm{Joergen W.} \spfx{van der} \sur{Ploeg} 
%%  \sfx{IV}}\email{iauthor@gmail.com}
%%=============================================================%%

\author[1]{\fnm{Bin} \sur{Cheng}}\email{chengbin2422@mails.jlu.edu.cn}

\author*[2]{\fnm{Jiaxuan} \sur{Lu}}\email{lujiaxuan@pjlab.org.cn}
% \equalcont{These authors contributed equally to this work.}

% \affil*[1]{\orgdiv{College of Software}, \orgname{Jilin University}, \orgaddress{\street{No.2699 QianjinStreet}, \city{Changchun}, \postcode{130012}, \state{Jilin}, \country{P.R.China}}}

\affil[1]{\orgdiv{College of Software}, \orgname{Jilin University}, \orgaddress{\city{Changchun}, \postcode{130012}, \state{Jilin}, \country{China}}}

\affil[2]{\orgname{Shanghai AI Lab}, \orgaddress{\city{Xuhui District}, \postcode{200232}, \state{Shanghai}, \country{China}}}

%%==================================%%
%% Sample for unstructured abstract %%
%%==================================%%

\abstract{With the advent of the era of foundation models, pre-training and fine-tuning have become common paradigms. Recently, parameter-efficient fine-tuning has garnered widespread attention due to its better balance between the number of learnable parameters and performance. However, some current parameter-efficient fine-tuning methods only model a single modality and lack the utilization of structural knowledge in downstream tasks. To address this issue, this paper proposes a multi-modal parameter-efficient fine-tuning method based on graph networks. Each image is fed into a multi-modal large language model (MLLM) to generate a text description. The image and its corresponding text description are then processed by a frozen image encoder and text encoder to generate image features and text features, respectively. A graph is constructed based on the similarity of the multi-modal feature nodes, and knowledge and relationships relevant to these features are extracted from each node. Additionally, Elastic Weight Consolidation (EWC) regularization is incorporated into the loss function to mitigate the problem of forgetting during task learning. The proposed model achieves test accuracies on the OxfordPets, Flowers102, and Food101 datasets that improve by 4.45\%, 2.92\%, and 0.23\%, respectively. The code is available at https://github.com/yunche0/GA-Net/tree/master.}


%%\pacs[JEL Classification]{D8, H51}

%%\pacs[MSC Classification]{35A01, 65L10, 65L12, 65L20, 65L70}

\maketitle

\section{Introduction}\label{sec1}

With the onset of the era of foundational models, we have gradually entered a paradigm of pre-training and fine-tuning. In terms of downstream task adaptation, full fine-tuning requires adjusting all the parameters of a model to adapt to downstream tasks. However, as the scale of models and the number of tasks increase, such a method becomes inefficient. Consequently, numerous studies have focused on parameter-efficient fine-tuning, exploring strategies to efficiently adapt existing foundation models to downstream tasks. Previous parameter-efficient fine-tuning methods can be mainly categorized into three approaches. The first is prompt tuning \cite{vu2021spot}\cite{lu2024pathotune}, which aims to achieve fine-tuning by modifying the model input rather than the model structure. The second is prefix tuning \cite{li2021prefix}, which involves updating only task-specific trainable parameters within each layer. The third is adapter tuning \cite{houlsby2019parameter}\cite{chen2022adaptformer}, which achieves parameter-efficient fine-tuning by inserting adapter modules with a bottleneck architecture between layers.

In the realm of multi-modal learning, parameter-efficient fine-tuning methods have gained attention in recent research. Prompt vectors are used to align multi-modal data, achieving efficient multi-modal fusion in low-resource settings \cite{lu2022prompt}. The main idea of CoOp \cite{zhou2022learning} is to automatically design prompt texts. It keeps the pre-training parameters unchanged and uses a small amount of data to learn the appropriate cues. The TaskRes \cite{yu2023task} method directly adjusts the weights of text-based classifiers without requiring extensive prompt updates to text encoders or elaborate adapters. $\pi$-Tuning optimizes parameters by predicting and interpolating task similarity across visual, language, and visual-language tasks, achieving efficient cross-task transfer learning \cite{wu2023pi}. However, these works do not consider modeling the complex associations between modalities.

With respect to modeling complex associations for parameter-efficient fine-tuning, methods related to graph neural networks (GNNs) have been explored. The concept of timely tuning has been introduced \cite{zhu2023dynamic}. In the molecular domain, MolCPT \cite{diao2022molcpt} enhances graph embeddings by encoding additional molecular motif information. However, these works primarily focus on how to perform parameter-efficient fine-tuning for purely graph structures, without applying them to language-image multi-modal modeling.

Therefore, we propose a framework that combines graph structures with multi-modal parameter-efficient fine-tuning methods, enabling the learning of multi-modal information while considering the complex associations between modalities. The proposed model comprises four main modules: Multi-Modal Feature Extraction, Multi-Modal Graph Construction, Graph Adapter Net (GA-Net), and Prediction. In the Multi-Modal Feature Extraction module, each image is processed by a pre-trained MLLM model to obtain a text description for each image. The image and its corresponding text description are then processed by frozen image and text encoders to generate image features and text features, respectively. These features are combined into multi-modal features through feature concatenation. In the Multi-Modal Graph Construction module, a graph is constructed based on the similarity of multi-modal feature nodes. The GA-Net module then mines suitable knowledge from the graph nodes, resulting in features that have fully learned both textual and image information while considering their adjacency relationships. In the Prediction module, the loss function incorporates cross-entropy and EWC regularization \cite{aich2021elastic}, which can mitigate the forgetting problem in task learning.

Compared with the current SOTA method, the proposed method improves by 4.45\% in the OxfordPets dataset, 2.92\% in the Flowers102 dataset and 0.23\% in the Food101 dataset.

The contributions of this paper are summarized as follows:
\begin{itemize}
    \item A parameter-efficient fine-tuning method based on graph networks, GA-Net, is proposed. The proposed method combines graph structures with multi-modal parameter-efficient fine-tuning, learning both textual and image information while considering the adjacency relationships between different tokens.
    \item EWC regularization is introduced into the loss function. By incorporating the correlation between parameter importance and the loss function, the EWC loss effectively retains knowledge from previous tasks and reduces interference with previous tasks when learning new tasks, thereby alleviating the problem of forgetting in task learning.
    \item Compared to the SOTA model, the proposed model improves the test accuracy by 4.45\% on the OxfordPets dataset, 2.92\% on the Flowers102 dataset and 0.23\% on the Food101 dataset.
\end{itemize}

\section{Related Work}\label{sec2}

\textbf{Parameter-Efficient Fine-Tuning Methods (PEFT).} Full fine-tuning involves modifying all the model's parameters to suit downstream tasks. Yet, as models grow in scale and the number of tasks expands, this approach becomes increasingly inefficient. To address this issue, in recent years, the natural language processing (NLP) community has explored parameter-efficient fine-tuning techniques (PEFT) \cite{liao2023parameter}\cite{li2023prefix}\cite{gheini2022know}\cite{pouramini2024matching}\cite{li2024eftnet}. These techniques only require adjusting a small subset of parameters, thereby improving efficiency \cite{ding2023parameter}. For example, prompt tuning methods \cite{vu2021spot} attempt to achieve fine-tuning by modifying the model input rather than the model structure. Prefix tuning \cite{li2021prefix} updates only task-specific trainable parameters in each layer. Adapter tuning \cite{houlsby2019parameter}\cite{chen2022adaptformer} inserts adapter modules with bottleneck architectures between layers to achieve parameter-efficient fine-tuning. Additionally, methods like BitFit \cite{zaken2021bitfit} update only the bias terms and freeze the remaining parameters, while LoRA \cite{hu2021lora} reduces the number of trainable parameters by decomposing the weight matrix into low-rank matrices.

In multi-modal learning, parameter-efficient fine-tuning methods have gained widespread attention in recent research \cite{dutta2023performance}\cite{lialin2023scaling}\cite{lian2022scaling}\cite{hirano2022parameter}\cite{wang2023non}. Using prompt vectors to align multi-modal data achieves efficient multi-modal fusion in low-resource environments, excelling in tasks involving two or more data modalities \cite{lu2022prompt}. Research on scaling large multi-modal models (such as LLaVA and MiniGPT-4) has shown that parameter-efficient training methods like LoRA/QLoRA perform well in both multi-modal and language tasks, with performance comparable to full-model fine-tuning\cite{lu2023empirical}. The main idea of CoOp \cite{zhou2022learning} is to automatically design prompt texts. It keeps the pre-trained parameters unchanged and learns suitable prompts using a small amount of data. CLIP-adapter \cite{gao2024clip} inserts a random learnable module into the middle of the model. By updating this module, it better adapts to downstream tasks. Unlike CLIP-Adapter, Tip-Adapter \cite{zhang2022tip} does not require SGD to train the adapter. Instead, it constructs a query-key cache model from few-shot supervision to obtain the adapter's weights. TaskRes \cite{yu2023task} directly adjusts the weights of the text-based classifier without needing extensive prompt updates to the text encoder or carefully designed adapters. $\pi$-Tuning optimizes parameters by predicting and interpolating task similarity across visual, language, and visual-language tasks, achieving efficient cross-task transfer learning, especially effective in data-scarce situations \cite{wu2023pi}. PMF significantly reduces training memory usage by adding prompt vectors only in the deeper layers of a single-modal Transformer \cite{xu2023source}. However, these methods do not model the complex associations between modalities.

\textbf{Graph Neural Networks (GNN).} Early work by Scarselli et al. \cite{scarselli2008graph} laid the foundation by proposing a framework for learning node representations in graphs, capable of capturing dependencies between nodes through iterative information passing. The pioneering work by Bruna et al. \cite{chen2017supervised} applied convolutional neural networks to graph data in the spectral domain. However, this approach faced challenges in computational efficiency and generalization across different graph structures. The graph convolutional networks (GCNs) proposed by Kipf and Welling \cite{schlichtkrull2018modeling} are among the most influential models, performing convolution by aggregating feature information from the local neighborhood of nodes, thus achieving efficient and scalable learning. The graph attention networks (GATs) introduced by Veličković et al. \cite{velivckovic2017graph} employ an attention mechanism to weight the importance of neighboring nodes, allowing for more flexible and dynamic information aggregation. Additionally, GraphSAGE by Hamilton et al. \cite{hamilton2017inductive} introduced a sampling-based approach for large-scale graph representation learning. Numerous prior works have applied GNNs to association modeling tasks across various domains, including vision-based\cite{lu2023exploring}\cite{gao2024hypergraph}, text-based\cite{dai2022graph}\cite{zhang2023improving}, and graph-based\cite{chen2017supervised}\cite{schlichtkrull2018modeling}\cite{zhu2023dynamic}\cite{han2023temporal} applications.

Parameter-efficient tuning methods have also seen some exploration in the GNN domain. The concept of timely tuning has been applied \cite{zhu2023dynamic}. Although methods like GPF \cite{meng2022regeneration} and GraphPrompt \cite{liu2023graphprompt} are also parameter-efficient, they struggle to match the benchmarks set by full fine-tuning in non-few-shot settings. GPPT \cite{sun2022gppt} designed a specific framework for GNNs, but its application is limited to node-level tasks. In the molecular domain, MolCPT \cite{diao2022molcpt} enhances graph embeddings by encoding additional molecular motif information. However, the aforementioned works design parameter-efficient fine-tuning methods specifically for pure graph structures. To the best of our knowledge, graph networks have not been applied in the field of multi-modal parameter-efficient fine-tuning.

\section{Method}\label{sec3}

\begin{figure}[htbp]
    \centering
    \includegraphics[width=1\textwidth,trim=48 570 30 49,clip]{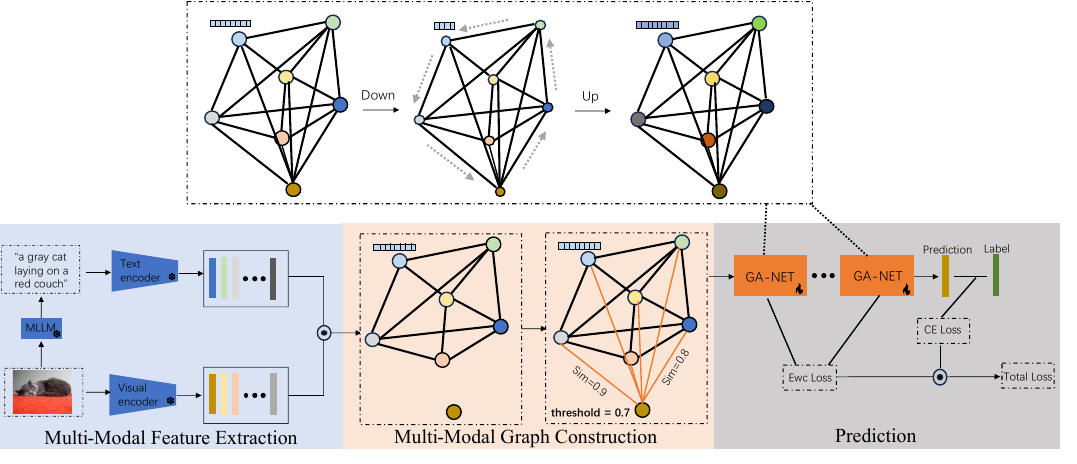}
    \caption{Overall pipeline of the proposed framework. The proposed model consists of four main modules: Multi-Modal Feature Extraction, Multi-Modal Graph Construction, GA-Net, and Prediction. GA-Net updates vertex features through down, GCN, and up operations, while the Prediction module uses a combined EWC and cross-entropy loss function to improve performance.}
    \label{fig:pic1}
\end{figure}

As shown in Figure \ref{fig:pic1}, the model presented in this paper is composed of four main modules: Multi-Modal Feature Extraction, Multi-Modal Graph Construction, GA-Net, and Prediction. The primary function of the Multi-Modal Feature Extraction is to use pre-trained models to extract features from images and text. Then, through Multi-Modal Graph Construction, a multi-modal graph is built to relate the connections between different modalities. The proposed GA-Net is the only trainable part of the network, where a down projection is first introduced, followed by a GCN, and finally an up projection to propagate and update the vertex features in the graph. Lastly, in the Prediction module, we propose a combined loss function of EWC loss and cross-entropy loss to further enhance the network's performance. The details of each module are as follows:

%\begin{figure}[htbp]
%    \centering
%    \includegraphics[width=1\textwidth,trim=30 600 80 30,clip]{pic1.pdf}
%    \caption{Overall pipeline of the proposed framework. The proposed model consists of four main modules: Multi-Modal Feature Extraction, Multi-Modal Graph Construction, GA-Net, and Prediction. GA-Net updates vertex features through down, GCN, and up operations, while the Prediction module uses a combined EWC and cross-entropy loss function to improve performance.}
%    \label{fig:pic1}
%\end{figure}

\subsection{Multi-Modal Feature Extraction}

In this module, we first use a pre-trained MLLM model to generate general text descriptions corresponding to the images. These descriptions do not involve the category names used in the final classification. Next, the images and their corresponding text descriptions are fed into frozen image encoders (such as ViT \cite{dosovitskiy2020image} or ResNet \cite{he2016deep}) and text encoders (such as BERT \cite{devlin2018bert}), respectively, to generate image features and text features. Finally, the image features and text features are combined into multi-modal features through feature concatenation. Mathematically, suppose $X_i$ is the input image, and $\text{MLLM}(\ast)$ represents the MLLM model, the text description of image $X_i$ obtained through the MLLM model can be expressed as
\begin{equation}
    X_i^t = \text{MLLM}(X_i)
\end{equation}

Let $I(X_i)$ and $T(X_i^t)$ represent the series of tokens obtained from the frozen image encoder (ViT/ResNet) and text encoder (BERT), respectively. These can be expressed as
\begin{equation}
    V_i^I = \{X_i^{I^1}, X_i^{I^2}, \ldots, X_i^{I^N}\} = I(X_i)
\end{equation}

\begin{equation}
    V_i^T = \{X_i^{T^1}, X_i^{T^2}, \ldots, X_i^{T^N}\} = T(X_i^t)
\end{equation}

where each token is of dimension $E$. The image tokens encode information from each patch location, while the text tokens encode information from each word location. $V_i^I$ and $V_i^T$ are sets of tokens. Each element in $V_i^I$ represents the $i$-th token in the image, and each element in $V_i^T$ represents the $i$-th token in the text. $N$ denotes the number of tokens. All tokens form a series of text and image vertices:
\begin{equation}
    V_i = \{V_i^I, V_i^T\} = \text{Concat}(V_i^I, V_i^T)
\end{equation}
where $\text{Concat}(\ast)$ represents feature concatenation, and $V_i$ represents a node after concatenating text features and image features.

\subsection{Multi-Modal Graph Construction}
To uncover the structural knowledge in the text embedding space for downstream tasks, i.e., the relationships between different semantics, and given the diverse visual features of different samples, we can measure finer-grained relationships between different semantics in the visual and textual space. Thus, we can construct a multi-modal graph structure \( G = \{V_i, E\} = \{V_i^I, V_i^T; E\} \), where \( V_i^I \) and \( V_i^T \) can be seen as the sets of image vertices and text vertices, respectively. \( E \) represents the set of edges.

We build the graph using the similarity of multi-modal features via a predefined threshold \(\gamma\). When the similarity between two multi-modal features is greater than \(\gamma\), an undirected edge is created between these two vertices, representing the adjacency relationships between all multi-modal features:
\begin{equation}
    E_{ij} =
    \begin{cases}
      1, & \text{if } i \neq j \text{ and } \text{Sim}(V_i, V_j) > \gamma \\
      0, & \text{otherwise}
    \end{cases}
\end{equation}
where
\begin{equation}
    \text{Sim}(V_i, V_j) = \frac{V_i \cdot V_j}{\|V_i\| \|V_j\|}
\end{equation}
represents the similarity between multi-modal nodes \( V_i \) and \( V_j \). \(\gamma\) is the similarity threshold, and an edge is constructed when the similarity between two vertices in the graph exceeds this threshold.

\subsection{Graph Adapter Net}
We propose a parameter-efficient fine-tuning method based on graph networks called Graph Adapter Net (GA-Net), where the rest of the network is frozen and only the GCN \cite{kipf2016semi} is fine-tuned during downstream tasks. The advantage of this method is that it can adapt to downstream tasks and improve model performance without significantly increasing the number of model parameters. Furthermore, since adapter fine-tuning only requires training a small number of parameters, it can significantly reduce computational resources and time costs for fine-tuning. The unique aspect of GA-Net is its ability to update features based on our constructed multi-modal graph structure, allowing fine-tuning while preserving adjacency relationships. The process of GA-Net can be represented by the following formula:
\begin{equation}
    C_{X_i} = W_{up} \left( \text{GCN} \left( W_{down}(X_i) \right) \right)
\end{equation}

\begin{figure}[htbp]
    \centering
    \includegraphics[width=1\textwidth,trim=33 670 40 30,clip]{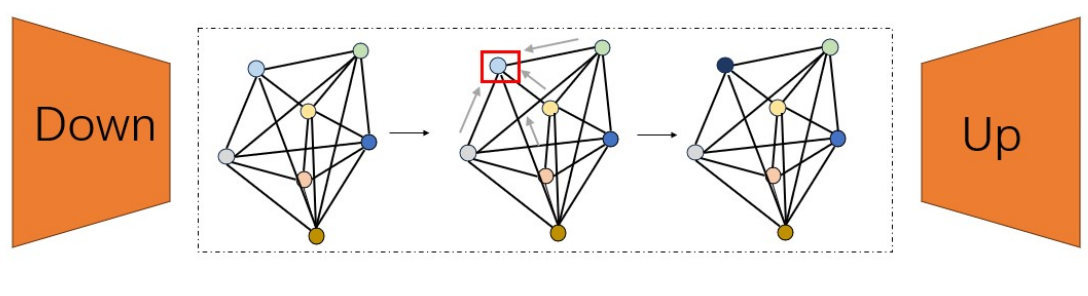}
    \caption{Details of GA-Net. By projecting the graph structure downwards, the number of model parameters is significantly reduced. Then, a GCN network aggregates information from adjacent nodes to learn complex associations between different modalities. Finally, the structure is projected upwards, restoring the original graph size.}
    \label{fig:pic2}
\end{figure}

where \( C_x \) represents the vertex feature matrix. The GA-Net utilizes a bottleneck architecture, including a down-projection \( W_{\text{down}}: \mathbb{R}^{n_{\text{in}}} \rightarrow \mathbb{R}^{n_{\text{mid}}} \), a Graph Convolutional Network (GCN) layer, and an up-projection \( W_{\text{up}}: \mathbb{R}^{n_{\text{mid}}} \rightarrow \mathbb{R}^{n_{\text{out}}} \). For vertex feature aggregation, we utilize the GCN, and the update formula for the vertex features in each layer can be formulated as:
\begin{equation}
    C^{(l+1)} = \sigma \left( \hat{D}^{-\frac{1}{2}} \hat{A} \hat{D}^{-\frac{1}{2}} C^{(l)} W^{(l)} \right)
\end{equation}
where \( \hat{A} \) is the normalized adjacency matrix, defined as \( \hat{A} = A + I \), where \( A \) is the original adjacency matrix and \( I \) is the identity matrix. \( \hat{D} \) is the degree matrix of \( \hat{A} \), with diagonal elements \( \hat{D}_{ii} = \sum_j \hat{A}_{ij} \). \( W^{(l)} \) is the trainable weight matrix of the \( l \)-th layer, with dimensions \( F_l \times F_{l+1} \). \( \sigma \) is the activation function, applied element-wise to the matrix. \( C^{(l)} \) represents the node feature matrix of the \( l \)-th layer of the GCN, with dimensions \( N \times F_l \), where \( N \) is the number of nodes and \( F_l \) is the feature dimension of the \( l \)-th layer. \( C^{(l+1)} \) represents the node feature matrix of the \( (l+1) \)-th layer of the GCN, with dimensions \( N \times F_{l+1} \), where \( F_{l+1} \) is the feature dimension of the \( (l+1) \)-th layer.

\subsection{Prediction}
In the prediction stage, we introduce Elastic Weight Consolidation (EWC) regularization \cite{aich2021elastic} into the generic cross-entropy loss function. By incorporating the importance of parameters and their association with the loss function, the EWC algorithm can effectively retain knowledge from previous tasks and reduce interference when learning new tasks, thereby addressing the problem of catastrophic forgetting in task learning.

First, we calculate the importance of parameters based on the training results of previous tasks, using the Fisher Information Matrix \( I_A \). The Fisher Information Matrix is calculated as follows:
\begin{equation}
    I_A = \mathbb{E}\left[\frac{\partial^2 \mathcal{L}(\theta)}{\partial \theta^2} \bigg|_{\theta^*}\right] = \mathbb{E}\left[\left(\frac{\partial \mathcal{L}(\theta)}{\partial \theta} \frac{\partial \mathcal{L}(\theta)}{\partial \theta^\top}\right) \bigg|_{\theta^*}\right]
\end{equation}
where \( I_A \) is the Fisher Information Matrix representing the importance of parameters; \( \mathbb{E} \) is the expectation operator; \( \mathcal{L}(\theta) \) is the cross-entropy loss function, representing the model's CE loss under parameters \( \theta \), with \( L_{CE} = \mathcal{L}(\theta) \); \( \theta \) are the model parameters; \( \theta^* \) are the optimal parameters obtained from previous task training; \( \frac{\partial^2 \mathcal{L}(\theta)}{\partial \theta^2} \) is the second derivative of the loss function with respect to the parameters, indicating curvature information; \( \frac{\partial \mathcal{L}(\theta)}{\partial \theta} \) is the first derivative of the loss function with respect to the parameters, indicating gradient information.

The EWC loss is defined as:
\begin{equation}
    L_{ewc} = \mathcal{L}_\mathcal{B}(\theta) + \frac{\lambda}{2} \sum_{i} I_A(i) \left(\theta_i - \theta_i^*\right)^2
\end{equation}
where \( \mathcal{L}_\mathcal{B}(\theta) \) is the loss function for the current task \( B \); \( \lambda \) is a hyperparameter that balances the learning of new tasks and the retention of old tasks; \( I_A(i) \) are the diagonal elements of the Fisher Information Matrix, representing the importance of parameter \( \theta_i \) in task \( A \); \( \theta_i^* \) are the parameters learned in task \( A \). The model, regularized by EWC, is used for the final prediction, improving performance on new tasks while mitigating the forgetting problem.

The total loss is the sum of the ordinary cross-entropy loss \( L_{CE} \) and the EWC loss \( L_{ewc} \):
\begin{equation}
    L_{total} = L_{CE} + L_{ewc}
\end{equation}

\section{Experiment}\label{sec4}
\subsection{Experiment Settings}
We validated our model on three downstream classification tasks: Oxford Pets \cite{parkhi2012cats}, Flowers102 \cite{nilsback2008automated}, and Food101 \cite{bossard2014food}. All these datasets belong to the fine-grained classification category. The Oxford Pets dataset contains 37 categories (25 dog breeds and 12 cat breeds) with a total of 7,349 images. The Flowers102 dataset includes 102 categories with a total of 8,189 images. The Food101 dataset consists of 101 food categories with a total of 101,000 images. These datasets are not only rich in categories but also possess high fine-grained characteristics, making them ideal for evaluating the model's performance in distinguishing similar categories.

\subsubsection{Implementation Details}
We use the LlaVA \cite{lin2023video} model to generate general text descriptions corresponding to the images, ensuring that these descriptions do not mention the category names for final classification. Unless otherwise stated, we use the pre-trained backbone ViT-B/16 \cite{dosovitskiy2020image} as the visual encoder to produce visual features. We optimized the model for 100 epochs. During training, we used a batch size of 16 and an Adam optimizer with an initial learning rate of \(1 \times 10^{-3}\), which decays following a cosine learning rate schedule.

\subsection{Comparisons with State-of-the-Arts}

\begin{table}[htbp]
    \centering
    \caption{Comparison with SOTA methods in terms of accuracy}
    \begin{tabular}{l|c|c|c}
    \hline
    \textbf{Method} & \textbf{Flowers102} & \textbf{Oxford Pets} & \textbf{Food101} \\
    \hline
    CLIP-Adapter   & 93.90 & 87.84 & 78.25 \\
    CoOp           & 94.51 & 87.01 & 74.67 \\
    TaskRes        & 96.10 & 88.10 & 78.23 \\
    Tip-Adapter    & 94.23 & 88.18 & 78.11 \\
    \textbf{Ours}  & \textbf{99.02} & \textbf{92.63} & \textbf{78.46} \\
    \hline
    \end{tabular}
    \label{tab:results}
\end{table}

We compared the proposed model with several state-of-the-art parameter-efficient fine-tuning methods, including TaskRes \cite{yu2023task}, CoOp \cite{zhou2022learning}, CLIP-adapter \cite{gao2024clip}, and Tip-Adapter \cite{zhang2022tip}, on the Oxford Pets, Flowers102, and Food101 datasets. As shown in Table \ref{tab:results}, the experimental results demonstrate that our model consistently outperforms previous parameter-efficient fine-tuning models on the average performance of the benchmark datasets. Our model achieved an average performance of 90.03\%, outperforming Tip-Adapter by 3.19\% and TaskRes by 2.82\%. The model's test accuracy improved by 4.45\% and 2.92\% on the Oxford Pets and Flowers102 datasets, respectively, compared to the state-of-the-art methods. Our model still performed the best on the Food101 dataset. The more significant improvement on the Oxford Pets and Food101 datasets is due to the higher need for multi-modal associations, which are better modeled through GNN. Similarly, Tip-Adapter performs better than other SOTA methods as it combines the strengths of prompts and adapters, introducing task-related prompts into the model to provide more multi-modal associations.

\subsection{Model Efficiency}
\begin{table}[htbp]
    \centering
    \caption{Comparison with SOTA methods in terms of efficiency}
    \begin{tabular}{l|c|c|c|c|c}
    \hline
    \textbf{Metric} & \textbf{CLIP-Adapter} & \textbf{CoOp} & \textbf{TaskRes} & \textbf{Tip-Adapter} & \textbf{Ours} \\
    \hline
    Parameters (M) & 0.524 & \textbf{0.008} & 1.024 & 16.384 & 0.056 \\
    Memory Cost (Training) & 9.257 & 18.907 & 6.227 & \textbf{4.313} & 13.826 \\
    Memory Cost (Inference) & 7.615 & 7.403 & 6.225 & \textbf{4.161} & 6.713 \\
    \hline
    \end{tabular}
    \label{tab:efficiency}
\end{table}

As shown in Table \ref{tab:efficiency}, we conducted experiments on the parameter quantities of different methods. The parameter consumption of Tip-Adapter is exceptionally high and is not in the same range as other methods. Among the remaining three models, TaskRes has a higher parameter count than CLIP-Adapter, with CoOp having the least parameter count. Our model has a parameter count of 0.056M, which is less than most models and only slightly higher than CoOp. However, the accuracy of our model significantly surpasses that of CoOp, indicating that our model achieves a good balance between accuracy and parameter count.

When the image encoder and text encoder are not frozen, the parameter size is 195,337,765. When the image encoder and text encoder are frozen, the parameter size is 56,869. Therefore, with only 0.029\% of the parameters being trainable, our model can surpass the state-of-the-art (SOTA) models on multiple datasets. Our model ranks second in terms of model efficiency among SOTA models. This is partly because the text descriptions in the CoOp model are generated using fixed handcrafted sentences, which are relatively simple text descriptions. Compared to our model, even though the CoOp model has fewer training parameters, the quality of the textual information is lower, resulting in significantly lower accuracy on two datasets.

\subsection{Ablation Study}

\begin{table}[htbp]
\centering
\caption{The accuracy under different components}
\begin{tabular}{l|c|c|c|c}
\hline
\textbf{Baseline} & \textbf{EWC} & \textbf{GNN} & \textbf{Multi-Modal} & \textbf{Accuracy} \\
\hline
Baseline & \xmark & \xmark & \xmark & 86.20\% \\
Baseline-EWC & \checkmark & \xmark & \xmark & 87.85\% \\
Multi-Modal Baseline-EWC & \checkmark & \checkmark & \xmark & 90.08\% \\
GA-Net & \checkmark & \checkmark & \checkmark & 92.63\% \\
\hline
\end{tabular}
\label{tab:accuracy_components}
\end{table}

As shown in Table \ref{tab:accuracy_components}, the baseline is a simple linear layer trained on single-modal features, which are also extracted using a pre-trained model. Without using our foundational modules, the accuracy is only 86.20\%. After applying EWC regularization, the accuracy improves by 1.65\% to 87.85\%. When using multi-modal learning, the performance further increases by 2.23\% to 90.08\%. The improvement with multi-modal learning is because leveraging two modalities simultaneously provides greater capability compared to using a single modality. Finally, by integrating the complete graph method, the accuracy improves by 2.55\% to 92.23\%. The performance boost from incorporating the graph is due to its ability to better model the relationships between tokens, thereby demonstrating the effectiveness of our proposed method.

The text descriptions for each image are generated by MLLM. In the baseline experiments, replacing all text descriptions uniformly with "A photo of a pet/flower" eliminates text information during training. Ablation experiments show that different combinations of methods and features enhance the model's performance to varying degrees. The GA-Net model achieves the best performance by combining EWC\cite{aich2021elastic}, multi-modal features, and GNN. Introducing EWC regularization improves model performance by 1.65\%; introducing GNN enhances performance by 2.55\%; and incorporating multi-modal learning boosts performance by 2.23\%. The significant performance improvements from multi-modal learning and GNN indicate that our model can effectively capture complex associations between modalities. EWC regularization helps mitigating the forgetting problem in task learning, making its effect more pronounced with larger datasets.

In terms of memory cost, Tip-Adapter has the lowest memory consumption, while CoOp has the highest. This is contrary to the parameter usage, indicating that, to some extent, a greater number of parameters can lead to lower memory consumption. Although our model also incurs a significant memory cost, it is still less than CoOp and performs better. While Tip-Adapter consumes less memory than our model, our model requires far fewer parameters and delivers significantly better performance. Furthermore, our model needs to store adjacency relations between different tokens, which inevitably consumes some memory space. This aspect contributes to the performance of our model. This demonstrates that our model achieves a good balance between accuracy and parameter consumption.

\subsection{Hyperparameter Study}

\begin{figure}[htbp]
    \centering
    \includegraphics[width=0.8\textwidth]{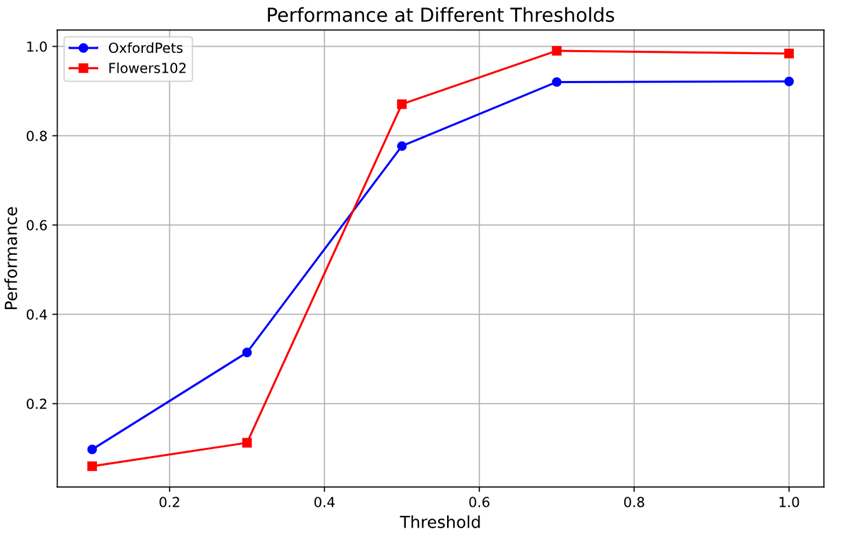}
    \caption{Accuracy under different similarity thresholds}
    \label{fig:pic3}
\end{figure}

To investigate the impact of hyperparameters, specifically similarity thresholds, we analyzed different similarity thresholds on the OxfordPets and Flowers102 datasets, as shown in Figure~\ref{fig:pic3}. The results indicate that different datasets are affected differently by varying similarity thresholds. For both datasets, the accuracy increased most significantly within the threshold range of 0.3 to 0.5. Additionally, the accuracy improvement for the Flowers102 dataset was more pronounced within this range compared to the OxfordPets dataset, suggesting that the Flowers102 dataset is more sensitive to adjacency relations in the graph. When the similarity threshold reaches 0.7, the model achieves peak accuracy on both datasets. For thresholds less than 0.7, accuracy shows an increasing trend with rising similarity thresholds. However, once the threshold exceeds 0.7, accuracy changes minimally, but the training burden increases, indicating that a similarity threshold of 0.7 is optimal.

\section{Conclusions}

This paper comprehensively reviews the limitations of previous parameter-efficient fine-tuning methods in low-data environments. These methods only model a single modality and lack the utilization of structural knowledge in downstream tasks. Therefore, we propose a novel parameter-efficient fine-tuning model, GA-Net, which extracts knowledge suitable for features from each multi-modal feature node in the graph, resulting in features that have fully learned both textual and image information while considering their adjacency relationships. Experiments on three fine-grained classification tasks, Oxford Pets, Flowers102, and Food101, demonstrate that the GA-Net model is effective in parameter-efficient fine-tuning.\\
The limitations of the model stem from the generation of text descriptions. In this paper, we use MLLM to generate text descriptions for each image. However, these prompts are simple and lack sufficient diversity. We believe that providing more diverse and accurate prompts for downstream tasks, such as using refined image caption methods, would better model the textual structural knowledge and further improve the performance of GA-Net.

\section{Declarations}\label{sec6}

\begin{itemize}
    \item \textbf{Funding}
    
    This research received no external funding.

    \item \textbf{Conflict of Interest/Competing Interests}
    
    We declare that we have no competing interests.

    \item \textbf{Ethics Approval and Consent to Participate}
    
    This study did not involve any human or animal subjects, hence ethical approval and consent to participate are not applicable.

    \item \textbf{Consent for Publication}
    
    We have reviewed the manuscript and consent to its publication.

    \item \textbf{Data Availability}
    
    The datasets used in this study are all publicly available:
    \begin{itemize}
        \item Oxford Pets dataset: \url{https://www.robots.ox.ac.uk/~vgg/data/pets/}
        \item Flowers-102 dataset: \url{https://www.robots.ox.ac.uk/~vgg/data/flowers/102/102flowers.tgz}
        \item Food101 dataset: \url{https://www.kaggle.com/datasets/dansbecker/food-101}
    \end{itemize}

    \item \textbf{Materials Availability}
    
    Not applicable.

    \item \textbf{Code Availability}
    
    The code used in this study is available at: \url{https://github.com/yunche0/GA-Net/tree/master}

    \item \textbf{Author Contributions}
    
    We all contributed significantly to the research and manuscript preparation, including the conception and design of the study, data collection and analysis, and manuscript writing. We have read and approved the final version of the manuscript.
\end{itemize}

%%===========================================================================================%%
%% If you are submitting to one of the Nature Portfolio journals, using the eJP submission   %%
%% system, please include the references within the manuscript file itself. You may do this  %%
%% by copying the reference list from your .bbl file, paste it into the main manuscript .tex %%
%% file, and delete the associated \verb+\bibliography+ commands.                            %%
%%===========================================================================================%%

\bibliography{sn-bibliography}% common bib file

%% BioMed_Central_Bib_Style_v1.01

\begin{thebibliography}{50}
% BibTex style file: bmc-mathphys.bst (version 2.1), 2014-07-24
\ifx \bisbn   \undefined \def \bisbn  #1{ISBN #1}\fi
\ifx \binits  \undefined \def \binits#1{#1}\fi
\ifx \bauthor  \undefined \def \bauthor#1{#1}\fi
\ifx \batitle  \undefined \def \batitle#1{#1}\fi
\ifx \bjtitle  \undefined \def \bjtitle#1{#1}\fi
\ifx \bvolume  \undefined \def \bvolume#1{\textbf{#1}}\fi
\ifx \byear  \undefined \def \byear#1{#1}\fi
\ifx \bissue  \undefined \def \bissue#1{#1}\fi
\ifx \bfpage  \undefined \def \bfpage#1{#1}\fi
\ifx \blpage  \undefined \def \blpage #1{#1}\fi
\ifx \burl  \undefined \def \burl#1{\textsf{#1}}\fi
\ifx \doiurl  \undefined \def \doiurl#1{\url{https://doi.org/#1}}\fi
\ifx \betal  \undefined \def \betal{\textit{et al.}}\fi
\ifx \binstitute  \undefined \def \binstitute#1{#1}\fi
\ifx \binstitutionaled  \undefined \def \binstitutionaled#1{#1}\fi
\ifx \bctitle  \undefined \def \bctitle#1{#1}\fi
\ifx \beditor  \undefined \def \beditor#1{#1}\fi
\ifx \bpublisher  \undefined \def \bpublisher#1{#1}\fi
\ifx \bbtitle  \undefined \def \bbtitle#1{#1}\fi
\ifx \bedition  \undefined \def \bedition#1{#1}\fi
\ifx \bseriesno  \undefined \def \bseriesno#1{#1}\fi
\ifx \blocation  \undefined \def \blocation#1{#1}\fi
\ifx \bsertitle  \undefined \def \bsertitle#1{#1}\fi
\ifx \bsnm \undefined \def \bsnm#1{#1}\fi
\ifx \bsuffix \undefined \def \bsuffix#1{#1}\fi
\ifx \bparticle \undefined \def \bparticle#1{#1}\fi
\ifx \barticle \undefined \def \barticle#1{#1}\fi
\bibcommenthead
\ifx \bconfdate \undefined \def \bconfdate #1{#1}\fi
\ifx \botherref \undefined \def \botherref #1{#1}\fi
\ifx \url \undefined \def \url#1{\textsf{#1}}\fi
\ifx \bchapter \undefined \def \bchapter#1{#1}\fi
\ifx \bbook \undefined \def \bbook#1{#1}\fi
\ifx \bcomment \undefined \def \bcomment#1{#1}\fi
\ifx \oauthor \undefined \def \oauthor#1{#1}\fi
\ifx \citeauthoryear \undefined \def \citeauthoryear#1{#1}\fi
\ifx \endbibitem  \undefined \def \endbibitem {}\fi
\ifx \bconflocation  \undefined \def \bconflocation#1{#1}\fi
\ifx \arxivurl  \undefined \def \arxivurl#1{\textsf{#1}}\fi
\csname PreBibitemsHook\endcsname

%%% 1
\bibitem[\protect\citeauthoryear{Vu et~al.}{2021}]{vu2021spot}
\begin{botherref}
\oauthor{\bsnm{Vu}, \binits{T.}},
\oauthor{\bsnm{Lester}, \binits{B.}},
\oauthor{\bsnm{Constant}, \binits{N.}},
\oauthor{\bsnm{Al-Rfou}, \binits{R.}},
\oauthor{\bsnm{Cer}, \binits{D.}}:
Spot: Better frozen model adaptation through soft prompt transfer.
arXiv preprint arXiv:2110.07904
(2021)
\end{botherref}
\endbibitem

%%% 2
\bibitem[\protect\citeauthoryear{Lu et~al.}{2024}]{lu2024pathotune}
\begin{botherref}
\oauthor{\bsnm{Lu}, \binits{J.}},
\oauthor{\bsnm{Yan}, \binits{F.}},
\oauthor{\bsnm{Zhang}, \binits{X.}},
\oauthor{\bsnm{Gao}, \binits{Y.}},
\oauthor{\bsnm{Zhang}, \binits{S.}}:
Pathotune: Adapting visual foundation model to pathological specialists.
arXiv preprint arXiv:2403.16497
(2024)
\end{botherref}
\endbibitem

%%% 3
\bibitem[\protect\citeauthoryear{Li and Liang}{2021}]{li2021prefix}
\begin{botherref}
\oauthor{\bsnm{Li}, \binits{X.L.}},
\oauthor{\bsnm{Liang}, \binits{P.}}:
Prefix-tuning: Optimizing continuous prompts for generation.
arXiv preprint arXiv:2101.00190
(2021)
\end{botherref}
\endbibitem

%%% 4
\bibitem[\protect\citeauthoryear{Houlsby et~al.}{2019}]{houlsby2019parameter}
\begin{bchapter}
\bauthor{\bsnm{Houlsby}, \binits{N.}},
\bauthor{\bsnm{Giurgiu}, \binits{A.}},
\bauthor{\bsnm{Jastrzebski}, \binits{S.}},
\bauthor{\bsnm{Morrone}, \binits{B.}},
\bauthor{\bsnm{De~Laroussilhe}, \binits{Q.}},
\bauthor{\bsnm{Gesmundo}, \binits{A.}},
\bauthor{\bsnm{Attariyan}, \binits{M.}},
\bauthor{\bsnm{Gelly}, \binits{S.}}:
\bctitle{Parameter-efficient transfer learning for nlp}.
In: \bbtitle{International Conference on Machine Learning},
pp. \bfpage{2790}--\blpage{2799}
(\byear{2019}).
\bcomment{PMLR}
\end{bchapter}
\endbibitem

%%% 5
\bibitem[\protect\citeauthoryear{Chen et~al.}{2022}]{chen2022adaptformer}
\begin{barticle}
\bauthor{\bsnm{Chen}, \binits{S.}},
\bauthor{\bsnm{Ge}, \binits{C.}},
\bauthor{\bsnm{Tong}, \binits{Z.}},
\bauthor{\bsnm{Wang}, \binits{J.}},
\bauthor{\bsnm{Song}, \binits{Y.}},
\bauthor{\bsnm{Wang}, \binits{J.}},
\bauthor{\bsnm{Luo}, \binits{P.}}:
\batitle{Adaptformer: Adapting vision transformers for scalable visual recognition}.
\bjtitle{Advances in Neural Information Processing Systems}
\bvolume{35},
\bfpage{16664}--\blpage{16678}
(\byear{2022})
\end{barticle}
\endbibitem

%%% 6
\bibitem[\protect\citeauthoryear{Lu et~al.}{2022}]{lu2022prompt}
\begin{bchapter}
\bauthor{\bsnm{Lu}, \binits{Y.}},
\bauthor{\bsnm{Liu}, \binits{J.}},
\bauthor{\bsnm{Zhang}, \binits{Y.}},
\bauthor{\bsnm{Liu}, \binits{Y.}},
\bauthor{\bsnm{Tian}, \binits{X.}}:
\bctitle{Prompt distribution learning}.
In: \bbtitle{Proceedings of the IEEE/CVF Conference on Computer Vision and Pattern Recognition},
pp. \bfpage{5206}--\blpage{5215}
(\byear{2022})
\end{bchapter}
\endbibitem

%%% 7
\bibitem[\protect\citeauthoryear{Zhou et~al.}{2022}]{zhou2022learning}
\begin{barticle}
\bauthor{\bsnm{Zhou}, \binits{K.}},
\bauthor{\bsnm{Yang}, \binits{J.}},
\bauthor{\bsnm{Loy}, \binits{C.C.}},
\bauthor{\bsnm{Liu}, \binits{Z.}}:
\batitle{Learning to prompt for vision-language models}.
\bjtitle{International Journal of Computer Vision}
\bvolume{130}(\bissue{9}),
\bfpage{2337}--\blpage{2348}
(\byear{2022})
\end{barticle}
\endbibitem

%%% 8
\bibitem[\protect\citeauthoryear{Yu et~al.}{2023}]{yu2023task}
\begin{bchapter}
\bauthor{\bsnm{Yu}, \binits{T.}},
\bauthor{\bsnm{Lu}, \binits{Z.}},
\bauthor{\bsnm{Jin}, \binits{X.}},
\bauthor{\bsnm{Chen}, \binits{Z.}},
\bauthor{\bsnm{Wang}, \binits{X.}}:
\bctitle{Task residual for tuning vision-language models}.
In: \bbtitle{Proceedings of the IEEE/CVF Conference on Computer Vision and Pattern Recognition},
pp. \bfpage{10899}--\blpage{10909}
(\byear{2023})
\end{bchapter}
\endbibitem

%%% 9
\bibitem[\protect\citeauthoryear{Wu et~al.}{2023}]{wu2023pi}
\begin{bchapter}
\bauthor{\bsnm{Wu}, \binits{C.}},
\bauthor{\bsnm{Wang}, \binits{T.}},
\bauthor{\bsnm{Ge}, \binits{Y.}},
\bauthor{\bsnm{Lu}, \binits{Z.}},
\bauthor{\bsnm{Zhou}, \binits{R.}},
\bauthor{\bsnm{Shan}, \binits{Y.}},
\bauthor{\bsnm{Luo}, \binits{P.}}:
\bctitle{$\pi$-tuning: Transferring multimodal foundation models with optimal multi-task interpolation}.
In: \bbtitle{International Conference on Machine Learning},
pp. \bfpage{37713}--\blpage{37727}
(\byear{2023}).
\bcomment{PMLR}
\end{bchapter}
\endbibitem

%%% 10
\bibitem[\protect\citeauthoryear{Zhu et~al.}{2023}]{zhu2023dynamic}
\begin{botherref}
\oauthor{\bsnm{Zhu}, \binits{X.}},
\oauthor{\bsnm{Wu}, \binits{Y.}},
\oauthor{\bsnm{Zhang}, \binits{Q.}},
\oauthor{\bsnm{Chen}, \binits{Z.}},
\oauthor{\bsnm{He}, \binits{Y.}}:
Dynamic link prediction for new nodes in temporal graph networks.
arXiv preprint arXiv:2310.09787
(2023)
\end{botherref}
\endbibitem

%%% 11
\bibitem[\protect\citeauthoryear{Diao et~al.}{2022}]{diao2022molcpt}
\begin{botherref}
\oauthor{\bsnm{Diao}, \binits{C.}},
\oauthor{\bsnm{Zhou}, \binits{K.}},
\oauthor{\bsnm{Liu}, \binits{Z.}},
\oauthor{\bsnm{Huang}, \binits{X.}},
\oauthor{\bsnm{Hu}, \binits{X.}}:
Molcpt: Molecule continuous prompt tuning to generalize molecular representation learning.
arXiv preprint arXiv:2212.10614
(2022)
\end{botherref}
\endbibitem

%%% 12
\bibitem[\protect\citeauthoryear{Aich}{2021}]{aich2021elastic}
\begin{botherref}
\oauthor{\bsnm{Aich}, \binits{A.}}:
Elastic weight consolidation (ewc): Nuts and bolts.
arXiv preprint arXiv:2105.04093
(2021)
\end{botherref}
\endbibitem

%%% 13
\bibitem[\protect\citeauthoryear{Liao et~al.}{2023}]{liao2023parameter}
\begin{botherref}
\oauthor{\bsnm{Liao}, \binits{B.}},
\oauthor{\bsnm{Meng}, \binits{Y.}},
\oauthor{\bsnm{Monz}, \binits{C.}}:
Parameter-efficient fine-tuning without introducing new latency.
arXiv preprint arXiv:2305.16742
(2023)
\end{botherref}
\endbibitem

%%% 14
\bibitem[\protect\citeauthoryear{Li et~al.}{2023}]{li2023prefix}
\begin{botherref}
\oauthor{\bsnm{Li}, \binits{J.}},
\oauthor{\bsnm{Aitken}, \binits{W.}},
\oauthor{\bsnm{Bhambhoria}, \binits{R.}},
\oauthor{\bsnm{Zhu}, \binits{X.}}:
Prefix propagation: Parameter-efficient tuning for long sequences.
arXiv preprint arXiv:2305.12086
(2023)
\end{botherref}
\endbibitem

%%% 15
\bibitem[\protect\citeauthoryear{Gheini et~al.}{2022}]{gheini2022know}
\begin{botherref}
\oauthor{\bsnm{Gheini}, \binits{M.}},
\oauthor{\bsnm{Ma}, \binits{X.}},
\oauthor{\bsnm{May}, \binits{J.}}:
Know where you're going: Meta-learning for parameter-efficient fine-tuning.
arXiv preprint arXiv:2205.12453
(2022)
\end{botherref}
\endbibitem

%%% 16
\bibitem[\protect\citeauthoryear{Pouramini and Faili}{2024}]{pouramini2024matching}
\begin{botherref}
\oauthor{\bsnm{Pouramini}, \binits{A.}},
\oauthor{\bsnm{Faili}, \binits{H.}}:
Matching tasks to objectives: Fine-tuning and prompt-tuning strategies for encoder-decoder pre-trained language models.
Applied Intelligence,
1--28
(2024)
\end{botherref}
\endbibitem

%%% 17
\bibitem[\protect\citeauthoryear{Li et~al.}{2024}]{li2024eftnet}
\begin{botherref}
\oauthor{\bsnm{Li}, \binits{J.}},
\oauthor{\bsnm{Wang}, \binits{Y.}},
\oauthor{\bsnm{Gao}, \binits{Z.}},
\oauthor{\bsnm{Wei}, \binits{Y.}}:
Eftnet: an efficient fine-tuning method for few-shot segmentation.
Applied Intelligence,
1--20
(2024)
\end{botherref}
\endbibitem

%%% 18
\bibitem[\protect\citeauthoryear{Ding et~al.}{2023}]{ding2023parameter}
\begin{barticle}
\bauthor{\bsnm{Ding}, \binits{N.}},
\bauthor{\bsnm{Qin}, \binits{Y.}},
\bauthor{\bsnm{Yang}, \binits{G.}},
\bauthor{\bsnm{Wei}, \binits{F.}},
\bauthor{\bsnm{Yang}, \binits{Z.}},
\bauthor{\bsnm{Su}, \binits{Y.}},
\bauthor{\bsnm{Hu}, \binits{S.}},
\bauthor{\bsnm{Chen}, \binits{Y.}},
\bauthor{\bsnm{Chan}, \binits{C.-M.}},
\bauthor{\bsnm{Chen}, \binits{W.}}, \betal:
\batitle{Parameter-efficient fine-tuning of large-scale pre-trained language models}.
\bjtitle{Nature Machine Intelligence}
\bvolume{5}(\bissue{3}),
\bfpage{220}--\blpage{235}
(\byear{2023})
\end{barticle}
\endbibitem

%%% 19
\bibitem[\protect\citeauthoryear{Zaken et~al.}{2021}]{zaken2021bitfit}
\begin{botherref}
\oauthor{\bsnm{Zaken}, \binits{E.B.}},
\oauthor{\bsnm{Ravfogel}, \binits{S.}},
\oauthor{\bsnm{Goldberg}, \binits{Y.}}:
Bitfit: Simple parameter-efficient fine-tuning for transformer-based masked language-models.
arXiv preprint arXiv:2106.10199
(2021)
\end{botherref}
\endbibitem

%%% 20
\bibitem[\protect\citeauthoryear{Hu et~al.}{2021}]{hu2021lora}
\begin{botherref}
\oauthor{\bsnm{Hu}, \binits{E.J.}},
\oauthor{\bsnm{Shen}, \binits{Y.}},
\oauthor{\bsnm{Wallis}, \binits{P.}},
\oauthor{\bsnm{Allen-Zhu}, \binits{Z.}},
\oauthor{\bsnm{Li}, \binits{Y.}},
\oauthor{\bsnm{Wang}, \binits{S.}},
\oauthor{\bsnm{Wang}, \binits{L.}},
\oauthor{\bsnm{Chen}, \binits{W.}}:
Lora: Low-rank adaptation of large language models.
arXiv preprint arXiv:2106.09685
(2021)
\end{botherref}
\endbibitem

%%% 21
\bibitem[\protect\citeauthoryear{Dutta et~al.}{2023}]{dutta2023performance}
\begin{bchapter}
\bauthor{\bsnm{Dutta}, \binits{A.}},
\bauthor{\bsnm{Alcaraz}, \binits{J.}},
\bauthor{\bsnm{TehraniJamsaz}, \binits{A.}},
\bauthor{\bsnm{Cesar}, \binits{E.}},
\bauthor{\bsnm{Sikora}, \binits{A.}},
\bauthor{\bsnm{Jannesari}, \binits{A.}}:
\bctitle{Performance optimization using multimodal modeling and heterogeneous gnn}.
In: \bbtitle{Proceedings of the 32nd International Symposium on High-Performance Parallel and Distributed Computing},
pp. \bfpage{45}--\blpage{57}
(\byear{2023})
\end{bchapter}
\endbibitem

%%% 22
\bibitem[\protect\citeauthoryear{Lialin et~al.}{2023}]{lialin2023scaling}
\begin{botherref}
\oauthor{\bsnm{Lialin}, \binits{V.}},
\oauthor{\bsnm{Deshpande}, \binits{V.}},
\oauthor{\bsnm{Rumshisky}, \binits{A.}}:
Scaling down to scale up: A guide to parameter-efficient fine-tuning.
arXiv preprint arXiv:2303.15647
(2023)
\end{botherref}
\endbibitem

%%% 23
\bibitem[\protect\citeauthoryear{Lian et~al.}{2022}]{lian2022scaling}
\begin{barticle}
\bauthor{\bsnm{Lian}, \binits{D.}},
\bauthor{\bsnm{Zhou}, \binits{D.}},
\bauthor{\bsnm{Feng}, \binits{J.}},
\bauthor{\bsnm{Wang}, \binits{X.}}:
\batitle{Scaling \& shifting your features: A new baseline for efficient model tuning}.
\bjtitle{Advances in Neural Information Processing Systems}
\bvolume{35},
\bfpage{109}--\blpage{123}
(\byear{2022})
\end{barticle}
\endbibitem

%%% 24
\bibitem[\protect\citeauthoryear{Hirano and Izumi}{2022}]{hirano2022parameter}
\begin{bchapter}
\bauthor{\bsnm{Hirano}, \binits{M.}},
\bauthor{\bsnm{Izumi}, \binits{K.}}:
\bctitle{Parameter tuning method for multi-agent simulation using reinforcement learning}.
In: \bbtitle{2022 9th International Conference on Behavioural and Social Computing (BESC)},
pp. \bfpage{1}--\blpage{7}
(\byear{2022}).
\bcomment{IEEE}
\end{bchapter}
\endbibitem

%%% 25
\bibitem[\protect\citeauthoryear{Wang et~al.}{2023}]{wang2023non}
\begin{botherref}
\oauthor{\bsnm{Wang}, \binits{Y.}},
\oauthor{\bsnm{Wu}, \binits{J.}},
\oauthor{\bsnm{Dabral}, \binits{T.}},
\oauthor{\bsnm{Zhang}, \binits{J.}},
\oauthor{\bsnm{Brown}, \binits{G.}},
\oauthor{\bsnm{Lu}, \binits{C.-T.}},
\oauthor{\bsnm{Liu}, \binits{F.}},
\oauthor{\bsnm{Liang}, \binits{Y.}},
\oauthor{\bsnm{Pang}, \binits{B.}},
\oauthor{\bsnm{Bendersky}, \binits{M.}}, et al.:
Non-intrusive adaptation: Input-centric parameter-efficient fine-tuning for versatile multimodal modeling.
arXiv preprint arXiv:2310.12100
(2023)
\end{botherref}
\endbibitem

%%% 26
\bibitem[\protect\citeauthoryear{Lu et~al.}{2023}]{lu2023empirical}
\begin{botherref}
\oauthor{\bsnm{Lu}, \binits{Y.}},
\oauthor{\bsnm{Li}, \binits{C.}},
\oauthor{\bsnm{Liu}, \binits{H.}},
\oauthor{\bsnm{Yang}, \binits{J.}},
\oauthor{\bsnm{Gao}, \binits{J.}},
\oauthor{\bsnm{Shen}, \binits{Y.}}:
An empirical study of scaling instruct-tuned large multimodal models.
arXiv preprint arXiv:2309.09958
(2023)
\end{botherref}
\endbibitem

%%% 27
\bibitem[\protect\citeauthoryear{Gao et~al.}{2024}]{gao2024clip}
\begin{barticle}
\bauthor{\bsnm{Gao}, \binits{P.}},
\bauthor{\bsnm{Geng}, \binits{S.}},
\bauthor{\bsnm{Zhang}, \binits{R.}},
\bauthor{\bsnm{Ma}, \binits{T.}},
\bauthor{\bsnm{Fang}, \binits{R.}},
\bauthor{\bsnm{Zhang}, \binits{Y.}},
\bauthor{\bsnm{Li}, \binits{H.}},
\bauthor{\bsnm{Qiao}, \binits{Y.}}:
\batitle{Clip-adapter: Better vision-language models with feature adapters}.
\bjtitle{International Journal of Computer Vision}
\bvolume{132}(\bissue{2}),
\bfpage{581}--\blpage{595}
(\byear{2024})
\end{barticle}
\endbibitem

%%% 28
\bibitem[\protect\citeauthoryear{Zhang et~al.}{2022}]{zhang2022tip}
\begin{bchapter}
\bauthor{\bsnm{Zhang}, \binits{R.}},
\bauthor{\bsnm{Zhang}, \binits{W.}},
\bauthor{\bsnm{Fang}, \binits{R.}},
\bauthor{\bsnm{Gao}, \binits{P.}},
\bauthor{\bsnm{Li}, \binits{K.}},
\bauthor{\bsnm{Dai}, \binits{J.}},
\bauthor{\bsnm{Qiao}, \binits{Y.}},
\bauthor{\bsnm{Li}, \binits{H.}}:
\bctitle{Tip-adapter: Training-free adaption of clip for few-shot classification}.
In: \bbtitle{European Conference on Computer Vision},
pp. \bfpage{493}--\blpage{510}
(\byear{2022}).
\bcomment{Springer}
\end{bchapter}
\endbibitem

%%% 29
\bibitem[\protect\citeauthoryear{Xu et~al.}{2023}]{xu2023source}
\begin{barticle}
\bauthor{\bsnm{Xu}, \binits{B.}},
\bauthor{\bsnm{Xu}, \binits{H.}},
\bauthor{\bsnm{Zhao}, \binits{H.}},
\bauthor{\bsnm{Gao}, \binits{J.}},
\bauthor{\bsnm{Liang}, \binits{D.}},
\bauthor{\bsnm{Li}, \binits{Y.}},
\bauthor{\bsnm{Wang}, \binits{W.}},
\bauthor{\bsnm{Feng}, \binits{Y.}},
\bauthor{\bsnm{Shi}, \binits{G.}}:
\batitle{Source apportionment of fine particulate matter at a megacity in china, using an improved regularization supervised pmf model}.
\bjtitle{Science of the Total Environment}
\bvolume{879},
\bfpage{163198}
(\byear{2023})
\end{barticle}
\endbibitem

%%% 30
\bibitem[\protect\citeauthoryear{Scarselli et~al.}{2008}]{scarselli2008graph}
\begin{barticle}
\bauthor{\bsnm{Scarselli}, \binits{F.}},
\bauthor{\bsnm{Gori}, \binits{M.}},
\bauthor{\bsnm{Tsoi}, \binits{A.C.}},
\bauthor{\bsnm{Hagenbuchner}, \binits{M.}},
\bauthor{\bsnm{Monfardini}, \binits{G.}}:
\batitle{The graph neural network model}.
\bjtitle{IEEE transactions on neural networks}
\bvolume{20}(\bissue{1}),
\bfpage{61}--\blpage{80}
(\byear{2008})
\end{barticle}
\endbibitem

%%% 31
\bibitem[\protect\citeauthoryear{Chen et~al.}{2017}]{chen2017supervised}
\begin{botherref}
\oauthor{\bsnm{Chen}, \binits{Z.}},
\oauthor{\bsnm{Li}, \binits{X.}},
\oauthor{\bsnm{Bruna}, \binits{J.}}:
Supervised community detection with line graph neural networks.
arXiv preprint arXiv:1705.08415
(2017)
\end{botherref}
\endbibitem

%%% 32
\bibitem[\protect\citeauthoryear{Schlichtkrull et~al.}{2018}]{schlichtkrull2018modeling}
\begin{bchapter}
\bauthor{\bsnm{Schlichtkrull}, \binits{M.}},
\bauthor{\bsnm{Kipf}, \binits{T.N.}},
\bauthor{\bsnm{Bloem}, \binits{P.}},
\bauthor{\bsnm{Van Den~Berg}, \binits{R.}},
\bauthor{\bsnm{Titov}, \binits{I.}},
\bauthor{\bsnm{Welling}, \binits{M.}}:
\bctitle{Modeling relational data with graph convolutional networks}.
In: \bbtitle{The Semantic Web: 15th International Conference, ESWC 2018, Heraklion, Crete, Greece, June 3--7, 2018, Proceedings 15},
pp. \bfpage{593}--\blpage{607}
(\byear{2018}).
\bcomment{Springer}
\end{bchapter}
\endbibitem

%%% 33
\bibitem[\protect\citeauthoryear{Veli{\v{c}}kovi{\'c} et~al.}{2017}]{velivckovic2017graph}
\begin{botherref}
\oauthor{\bsnm{Veli{\v{c}}kovi{\'c}}, \binits{P.}},
\oauthor{\bsnm{Cucurull}, \binits{G.}},
\oauthor{\bsnm{Casanova}, \binits{A.}},
\oauthor{\bsnm{Romero}, \binits{A.}},
\oauthor{\bsnm{Lio}, \binits{P.}},
\oauthor{\bsnm{Bengio}, \binits{Y.}}:
Graph attention networks.
arXiv preprint arXiv:1710.10903
(2017)
\end{botherref}
\endbibitem

%%% 34
\bibitem[\protect\citeauthoryear{Hamilton et~al.}{2017}]{hamilton2017inductive}
\begin{botherref}
\oauthor{\bsnm{Hamilton}, \binits{W.}},
\oauthor{\bsnm{Ying}, \binits{Z.}},
\oauthor{\bsnm{Leskovec}, \binits{J.}}:
Inductive representation learning on large graphs.
Advances in neural information processing systems
\textbf{30}
(2017)
\end{botherref}
\endbibitem

%%% 35
\bibitem[\protect\citeauthoryear{Lu et~al.}{2023}]{lu2023exploring}
\begin{barticle}
\bauthor{\bsnm{Lu}, \binits{J.}},
\bauthor{\bsnm{Wan}, \binits{H.}},
\bauthor{\bsnm{Li}, \binits{P.}},
\bauthor{\bsnm{Zhao}, \binits{X.}},
\bauthor{\bsnm{Ma}, \binits{N.}},
\bauthor{\bsnm{Gao}, \binits{Y.}}:
\batitle{Exploring high-order spatio--temporal correlations from skeleton for person re-identification}.
\bjtitle{IEEE Transactions on Image Processing}
\bvolume{32},
\bfpage{949}--\blpage{963}
(\byear{2023})
\end{barticle}
\endbibitem

%%% 36
\bibitem[\protect\citeauthoryear{Gao et~al.}{2024}]{gao2024hypergraph}
\begin{botherref}
\oauthor{\bsnm{Gao}, \binits{Y.}},
\oauthor{\bsnm{Lu}, \binits{J.}},
\oauthor{\bsnm{Li}, \binits{S.}},
\oauthor{\bsnm{Li}, \binits{Y.}},
\oauthor{\bsnm{Du}, \binits{S.}}:
Hypergraph-based multi-view action recognition using event cameras.
IEEE Transactions on Pattern Analysis and Machine Intelligence
(2024)
\end{botherref}
\endbibitem

%%% 37
\bibitem[\protect\citeauthoryear{Dai et~al.}{2022}]{dai2022graph}
\begin{barticle}
\bauthor{\bsnm{Dai}, \binits{Y.}},
\bauthor{\bsnm{Shou}, \binits{L.}},
\bauthor{\bsnm{Gong}, \binits{M.}},
\bauthor{\bsnm{Xia}, \binits{X.}},
\bauthor{\bsnm{Kang}, \binits{Z.}},
\bauthor{\bsnm{Xu}, \binits{Z.}},
\bauthor{\bsnm{Jiang}, \binits{D.}}:
\batitle{Graph fusion network for text classification}.
\bjtitle{Knowledge-based systems}
\bvolume{236},
\bfpage{107659}
(\byear{2022})
\end{barticle}
\endbibitem

%%% 38
\bibitem[\protect\citeauthoryear{Zhang et~al.}{2023}]{zhang2023improving}
\begin{barticle}
\bauthor{\bsnm{Zhang}, \binits{F.}},
\bauthor{\bsnm{Li}, \binits{J.}},
\bauthor{\bsnm{Cheng}, \binits{J.}}:
\batitle{Improving entity alignment via attribute and external knowledge filtering}.
\bjtitle{Applied Intelligence}
\bvolume{53}(\bissue{6}),
\bfpage{6671}--\blpage{6681}
(\byear{2023})
\end{barticle}
\endbibitem

%%% 39
\bibitem[\protect\citeauthoryear{Han et~al.}{2023}]{han2023temporal}
\begin{barticle}
\bauthor{\bsnm{Han}, \binits{D.}},
\bauthor{\bsnm{Kim}, \binits{D.}},
\bauthor{\bsnm{Kim}, \binits{M.}},
\bauthor{\bsnm{Han}, \binits{K.}},
\bauthor{\bsnm{Yi}, \binits{M.Y.}}:
\batitle{Temporal enhanced inductive graph knowledge tracing}.
\bjtitle{Applied Intelligence}
\bvolume{53}(\bissue{23}),
\bfpage{29282}--\blpage{29299}
(\byear{2023})
\end{barticle}
\endbibitem

%%% 40
\bibitem[\protect\citeauthoryear{Meng et~al.}{2022}]{meng2022regeneration}
\begin{barticle}
\bauthor{\bsnm{Meng}, \binits{Z.}},
\bauthor{\bsnm{Chen}, \binits{Z.}},
\bauthor{\bsnm{Tan}, \binits{J.}},
\bauthor{\bsnm{Wang}, \binits{W.}},
\bauthor{\bsnm{Zhang}, \binits{Z.}},
\bauthor{\bsnm{Huang}, \binits{J.}},
\bauthor{\bsnm{Fang}, \binits{J.}}:
\batitle{Regeneration performance and particulate emission characteristics during active regeneration process of gpf with ash loading}.
\bjtitle{Chemical Engineering Science}
\bvolume{248},
\bfpage{117114}
(\byear{2022})
\end{barticle}
\endbibitem

%%% 41
\bibitem[\protect\citeauthoryear{Liu et~al.}{2023}]{liu2023graphprompt}
\begin{bchapter}
\bauthor{\bsnm{Liu}, \binits{Z.}},
\bauthor{\bsnm{Yu}, \binits{X.}},
\bauthor{\bsnm{Fang}, \binits{Y.}},
\bauthor{\bsnm{Zhang}, \binits{X.}}:
\bctitle{Graphprompt: Unifying pre-training and downstream tasks for graph neural networks}.
In: \bbtitle{Proceedings of the ACM Web Conference 2023},
pp. \bfpage{417}--\blpage{428}
(\byear{2023})
\end{bchapter}
\endbibitem

%%% 42
\bibitem[\protect\citeauthoryear{Sun et~al.}{2022}]{sun2022gppt}
\begin{bchapter}
\bauthor{\bsnm{Sun}, \binits{M.}},
\bauthor{\bsnm{Zhou}, \binits{K.}},
\bauthor{\bsnm{He}, \binits{X.}},
\bauthor{\bsnm{Wang}, \binits{Y.}},
\bauthor{\bsnm{Wang}, \binits{X.}}:
\bctitle{Gppt: Graph pre-training and prompt tuning to generalize graph neural networks}.
In: \bbtitle{Proceedings of the 28th ACM SIGKDD Conference on Knowledge Discovery and Data Mining},
pp. \bfpage{1717}--\blpage{1727}
(\byear{2022})
\end{bchapter}
\endbibitem

%%% 43
\bibitem[\protect\citeauthoryear{Dosovitskiy et~al.}{2020}]{dosovitskiy2020image}
\begin{botherref}
\oauthor{\bsnm{Dosovitskiy}, \binits{A.}},
\oauthor{\bsnm{Beyer}, \binits{L.}},
\oauthor{\bsnm{Kolesnikov}, \binits{A.}},
\oauthor{\bsnm{Weissenborn}, \binits{D.}},
\oauthor{\bsnm{Zhai}, \binits{X.}},
\oauthor{\bsnm{Unterthiner}, \binits{T.}},
\oauthor{\bsnm{Dehghani}, \binits{M.}},
\oauthor{\bsnm{Minderer}, \binits{M.}},
\oauthor{\bsnm{Heigold}, \binits{G.}},
\oauthor{\bsnm{Gelly}, \binits{S.}}, et al.:
An image is worth 16x16 words: Transformers for image recognition at scale.
arXiv preprint arXiv:2010.11929
(2020)
\end{botherref}
\endbibitem

%%% 44
\bibitem[\protect\citeauthoryear{He et~al.}{2016}]{he2016deep}
\begin{bchapter}
\bauthor{\bsnm{He}, \binits{K.}},
\bauthor{\bsnm{Zhang}, \binits{X.}},
\bauthor{\bsnm{Ren}, \binits{S.}},
\bauthor{\bsnm{Sun}, \binits{J.}}:
\bctitle{Deep residual learning for image recognition}.
In: \bbtitle{Proceedings of the IEEE Conference on Computer Vision and Pattern Recognition},
pp. \bfpage{770}--\blpage{778}
(\byear{2016})
\end{bchapter}
\endbibitem

%%% 45
\bibitem[\protect\citeauthoryear{Devlin et~al.}{2018}]{devlin2018bert}
\begin{botherref}
\oauthor{\bsnm{Devlin}, \binits{J.}},
\oauthor{\bsnm{Chang}, \binits{M.-W.}},
\oauthor{\bsnm{Lee}, \binits{K.}},
\oauthor{\bsnm{Toutanova}, \binits{K.}}:
Bert: Pre-training of deep bidirectional transformers for language understanding.
arXiv preprint arXiv:1810.04805
(2018)
\end{botherref}
\endbibitem

%%% 46
\bibitem[\protect\citeauthoryear{Kipf and Welling}{2016}]{kipf2016semi}
\begin{botherref}
\oauthor{\bsnm{Kipf}, \binits{T.N.}},
\oauthor{\bsnm{Welling}, \binits{M.}}:
Semi-supervised classification with graph convolutional networks.
arXiv preprint arXiv:1609.02907
(2016)
\end{botherref}
\endbibitem

%%% 47
\bibitem[\protect\citeauthoryear{Parkhi et~al.}{2012}]{parkhi2012cats}
\begin{bchapter}
\bauthor{\bsnm{Parkhi}, \binits{O.M.}},
\bauthor{\bsnm{Vedaldi}, \binits{A.}},
\bauthor{\bsnm{Zisserman}, \binits{A.}},
\bauthor{\bsnm{Jawahar}, \binits{C.}}:
\bctitle{Cats and dogs}.
In: \bbtitle{2012 IEEE Conference on Computer Vision and Pattern Recognition},
pp. \bfpage{3498}--\blpage{3505}
(\byear{2012}).
\bcomment{IEEE}
\end{bchapter}
\endbibitem

%%% 48
\bibitem[\protect\citeauthoryear{Nilsback and Zisserman}{2008}]{nilsback2008automated}
\begin{bchapter}
\bauthor{\bsnm{Nilsback}, \binits{M.-E.}},
\bauthor{\bsnm{Zisserman}, \binits{A.}}:
\bctitle{Automated flower classification over a large number of classes}.
In: \bbtitle{2008 Sixth Indian Conference on Computer Vision, Graphics \& Image Processing},
pp. \bfpage{722}--\blpage{729}
(\byear{2008}).
\bcomment{IEEE}
\end{bchapter}
\endbibitem

%%% 49
\bibitem[\protect\citeauthoryear{Bossard et~al.}{2014}]{bossard2014food}
\begin{bchapter}
\bauthor{\bsnm{Bossard}, \binits{L.}},
\bauthor{\bsnm{Guillaumin}, \binits{M.}},
\bauthor{\bsnm{Van~Gool}, \binits{L.}}:
\bctitle{Food-101--mining discriminative components with random forests}.
In: \bbtitle{Computer vision--ECCV 2014: 13th European Conference, Zurich, Switzerland, September 6-12, 2014, Proceedings, Part VI 13},
pp. \bfpage{446}--\blpage{461}
(\byear{2014}).
\bcomment{Springer}
\end{bchapter}
\endbibitem

%%% 50
\bibitem[\protect\citeauthoryear{Lin et~al.}{2023}]{lin2023video}
\begin{botherref}
\oauthor{\bsnm{Lin}, \binits{B.}},
\oauthor{\bsnm{Zhu}, \binits{B.}},
\oauthor{\bsnm{Ye}, \binits{Y.}},
\oauthor{\bsnm{Ning}, \binits{M.}},
\oauthor{\bsnm{Jin}, \binits{P.}},
\oauthor{\bsnm{Yuan}, \binits{L.}}:
Video-llava: Learning united visual representation by alignment before projection.
arXiv preprint arXiv:2311.10122
(2023)
\end{botherref}
\endbibitem

\end{thebibliography}
%% if required, the content of .bbl file can be included here once bbl is generated
%%\input sn-article.bbl

\end{document}